\documentclass{article}
\pdfoutput=1

\usepackage{arxiv}

\usepackage[utf8]{inputenc} 
\usepackage[T1]{fontenc}    
\usepackage{hyperref}       
\usepackage{url}            
\usepackage{booktabs}       
\usepackage{amsfonts}       
\usepackage{nicefrac}       
\usepackage{microtype}      
\usepackage[pdftex]{graphicx}

\title{Using Natural Language Processing to Develop an Automated Orthodontic Diagnostic System}

\author{
  Tomoyuki Kajiwara \\
  Institute for Datability Science \\
  Osaka University \\
  Osaka, Japan \\
  \texttt{kajiwara@ids.osaka-u.ac.jp} \\
  \And Chihiro Tanikawa \\
  Graduate School of Dentistry \\
  Osaka University \\
  Osaka, Japan \\
  \texttt{ctanika@dent.osaka-u.ac.jp} \\
  \And Yuujin Shimizu \\
  Graduate School of Dentistry \\
  Osaka University \\
  Osaka, Japan \\
  \texttt{yjnshimizu@dent.osaka-u.ac.jp} \\
  \And Chenhui Chu \\
  Institute for Datability Science \\
  Osaka University \\
  Osaka, Japan \\
  \texttt{chu@ids.osaka-u.ac.jp} \\
  \And Takashi Yamashiro \\
  Graduate School of Dentistry \\
  Osaka University \\
  Osaka, Japan \\
  \texttt{yamashiro@dent.osaka-u.ac.jp} \\
  \And Hajime Nagahara \\
  Institute for Datability Science \\
  Osaka University \\
  Osaka, Japan \\
  \texttt{nagahara@ids.osaka-u.ac.jp} \\
}

\begin{document}
\maketitle

\begin{abstract}
We work on the task of automatically designing a treatment plan from the findings included in the medical certificate written by the dentist.
To develop an artificial intelligence system that deals with free-form certificates written by dentists, we annotate the findings and utilized the natural language processing approach.
As a result of the experiment using 990 certificates, $0.585$ F1-score was achieved for the task of extracting orthodontic problems from findings, and $0.584$ correlation coefficient with the human ranking was achieved for the treatment prioritization task.
\end{abstract}

\section{Introduction}
An orthodontic diagnosis and treatment plan involves predicting the entire course of action that a dentist should take to obtain the optimal treatment results at low risk~\cite{takada-2017}.
Making such an assessment – one that leads to the optimal diagnosis and treatment plan – requires years of knowledge and experience.
As such, there are cases in which inexperienced dentists make judgment errors or otherwise misunderstand a case's parameters.
An artificial intelligence (AI) system that can implement the years of experience of a specialist would be of great significance in providing patients with evidence-based medical care.
In addition, the automatic summarization of orthodontic diagnoses or presentation of necessary examinations in an orthodontic clinic would reduce the heavy workload of dentists, as well as help less experienced dentists in avoiding oversights and judgment errors.

Treatment operates on the logic of ``doing the opposite of the problem," and the process of modern orthodontic diagnosis and treatment consists mainly of the following three steps: $(1)$ Collection and itemization of patient information regarding the problem(s); $(2)$ Contemplating solutions for each problem; $(3)$ Determining the course of action and its implementation.
A certain regularity is apparent in the logical structure involved in medical diagnosis and treatment planning (as described above).
Hence, attempts have been made to automate orthodontic diagnosis and treatment planning, such as the use of expert systems.
An orthodontic diagnosis support system using fuzzy logic~\cite{simswilliams-1986}, as well as a support system for the selection of orthodontic appliances~\cite{stephens-1996}, have previously been developed.
However, orthodontic diagnosis and treatment planning support systems for use in dental practice are yet to be established.

To put the diagnosis process of a specialist in mathematical terms, if patient information regarding the problem is thought of as a set of feature values, the above step $(1)$ is analogous to representing medical conditions based on the individual weight of each feature value and detecting the degree of similarity between each medical condition.
Step $(2)$ can be considered equivalent to learning more about the approach to handle each of the medical conditions, at which point a natural language processing (NLP) AI system could be expected to find a solution.
The aim of this study is to develop an AI system that uses NLP on various imaging and model resources from the Graduate School and their accompanying treatment protocols.
Then, the findings are considered to not only create an automated process of diagnosis and treatment planning, but also put the results of the process in layman's terms for patient's ease of understanding.

\begin{figure}[t]
    \centering
    \includegraphics[width=\textwidth]{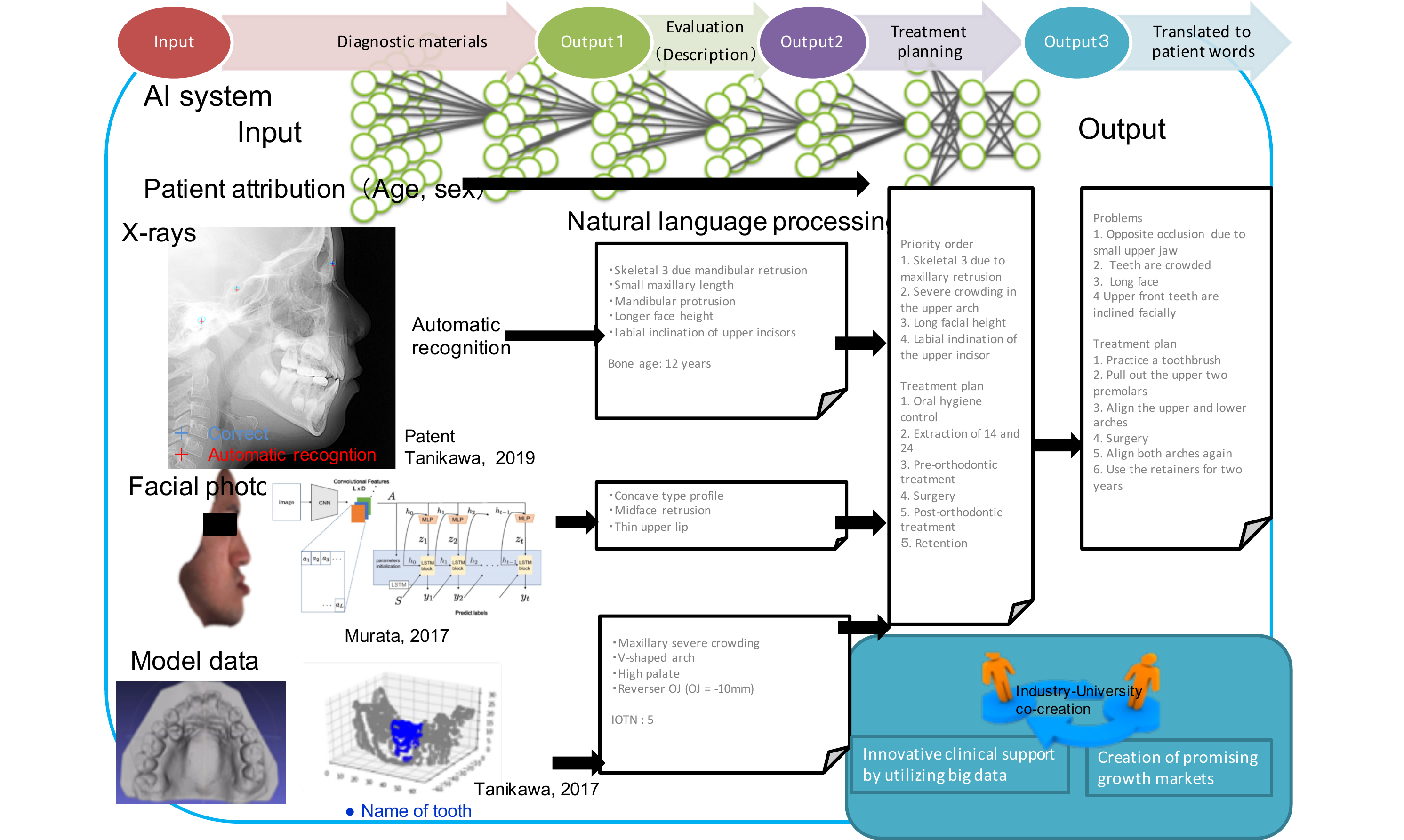}
    \caption{Automated diagnosis system in orthodontic treatment planning.}
    \label{fig:overview}
\end{figure}

\section{System Overview}
An overview of our system is shown in Figure~\ref{fig:overview}.
Based on the treatment protocol information (text data) kept by the Graduate School, the system uses NLP to extract imaging and model findings that represent a patient's medical condition.
This becomes the learning data.
Machine learning is then performed by extracting diagnoses (listed by priority) and treatment plans relevant to the findings and using them as answer data.

\subsection{Automated Summarization of Findings}

The system organizes each medical condition mentioned in the treatment protocol and formulates the automated treatment protocol summarization problem as a multi-label classification problem comprising of around $400$ types of medical condition labels.
Feature values are extracted from the text findings with NLP using vector space models, and the multi-label classification problem is solved using machine learning models, such as support vector machines (SVM) and deep learning models.
Furthermore, in the case of usable image findings, multi-modal feature extraction is jointly performed on the images to solve the multi-label classification problem with high accuracy.

\subsection{Automated Planning of Treatment Protocol}

Using the prioritized list of diagnoses results included in the treatment protocol, the system performs machine-learned ranking of each medical condition extracted at the automated summarization step.
Automated diagnosis is carried out by combining the medical condition extraction of Section 2.1 and the ranking of Section 2.2.

\subsection{Automated Simplification of Results for Reporting to a Patient}

The system trains a series transformation model using pairs of relevant treatment protocol summaries and consent form documents.
First, relevant sentence pairs are automatically extracted from the relevant documents using an NLP method of sentence alignment.
Next, statistical machine translation or neural machine translation techniques are used to automatically translate difficult text and generate text with simpler language that can be understood by patients easily.
This paper does not address text simplification. This is our future work.

\section{Dataset and Problem Setting}
\begin{figure}[t]
    \centering
    \includegraphics[width=0.95\textwidth]{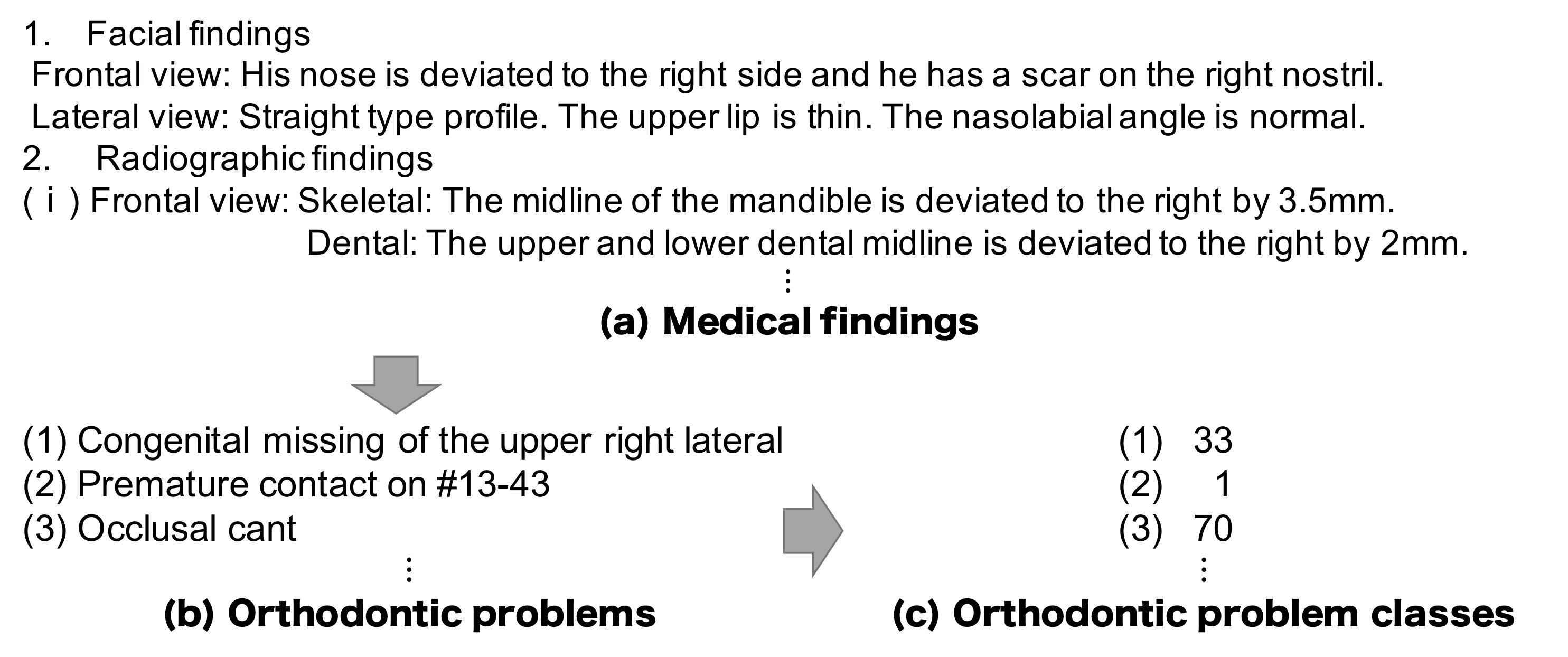}
    \caption{Annotation of orthodontic problem classes to the medical certificate.}
    \label{fig:annotation}
\end{figure}

In this study, we work on the task of automatically designing a treatment plan from the findings included in the medical certificate written by the dentist.
We used $990$ certificates written by dentists for our experiments.
We develop an artificial intelligence system that is input a document that describes the findings of the patient (Figure~\ref{fig:annotation} (a)) and outputs a list of problems that the patient has in the order of treatment priority (Figure~\ref{fig:annotation} (b)).
As findings are free-form descriptions in natural language, the method of feature extraction for machine learning is not obvious.
Therefore, we employ a natural language processing approach for feature extraction from text.

We divide into the following two subtasks to efficiently address the task of planning a treatment plan from findings:
\begin{itemize}
    \item Step~$1$: List the problems
    \item Step~$2$: Prioritize treatment
\end{itemize}
Step~$1$ summarizes the findings and lists the orthodontic problems each patient has.
This task can be regarded as a text generation task that generates a summary of findings.
However, compared to the millions to tens of millions of datasets commonly used in text generation tasks such as machine translation and automatic summarization in natural language processing, our dataset of $990$ documents is significantly smaller.
Therefore, we add an annotation as shown in Figure~\ref{fig:annotation} (c) and tackle Step~$1$ as a multi-label classification problem.
In this additional annotation, one dentist of the author classified orthodontic problems into $423$ classes.
Each problem that the patient has corresponds to one class.
Step~$2$ ranks each problem listed in Step~$1$ in terms of treatment priority.
Treatment priorities are indicated in brackets in Figures~\ref{fig:annotation} (b) and (c).
Solving the above two subtasks, our system automatically creates a treatment plan from the findings contained in the medical certificate.

According to the dataset, each patient has an average of $15.4$ orthodontic problems.
In the following experiments, these $990$ documents are randomly divided into $810$ for training, $90$ for validation, and $90$ for evaluation.

\section{Multi-label Classification Model for Orthodontic Problems}

\subsection{Proposed Methods}

We develop classification models of orthodontic problems on the datasets shown in (a) and (c) of Figure~\ref{fig:annotation}.
These models take the text of the findings as input and output a list of class labels in Figure~\ref{fig:annotation} (c).
By working as a classification problem rather than a generation problem, it can be expected that even small datasets will be less susceptible to linguistic variation.

We convert each text of findings into vector representation using a natural language processing approach and perform supervised learning of multi-label classification with them as features.
The following methods are used to vectorize each document.
\begin{itemize}
    \item BoW (Bag-of-Words): The BoW representations have dimensions corresponding to the number of vocabulary in the training dataset, and each dimension has a value of $1$ if the corresponding word appears in the input text or $0$ it does not. These are high-dimensional sparse vectors.
    \item USE (Universal Sentence Encoder): We construct feature vectors from the text using the cross-lingual version of universal sentence encoder~\cite{chidambaram-2018}. These are $512$-dimensional dense vectors.
\end{itemize}

A multi-layer perceptron was implemented on Chainer~\cite{tokui-2015} for a multi-label classifier.
In the output layer, a sigmoid function was used instead of the softmax function in the single-label classification.

\subsection{Experimental Settings}

Each sentence was divided into words using MeCab~\cite{kudo-2004} for the BoW model.
In this experiment, the vocabulary size is $2,075$ because only words appearing five or more times in the training dataset are used.
In addition, semantically equivalent classes are grouped, and $151$ class labels are used.
We used a three-layer perceptron for this experiment and examined the hidden layer size as a hyper-parameter from $\{256, 512, 1024, 2048, 4096\}$ on the validation dataset.

\subsection{Experimental Results}

The performance of each model was automatically evaluated using the F1-score which is the harmonic average of the precision and the recall.
The experimental results are shown in Table~\ref{tab:result_f}.
The simple BoW model achieved higher performance.
Unique structures such as bullets and incomplete sentences may have a negative impact on the sentence encoder.
On the other hand, BoW models treat documents as a set of words, so they are not affected by the structure of the sentence.

\begin{table}[t]
\begin{center}
\begin{tabular}{lccc}\toprule
    & Precision & Recall & F1-score \\ \midrule
BoW & 0.653     & 0.546  & 0.585    \\
USE & 0.634     & 0.486  & 0.536    \\ \bottomrule
\end{tabular}
\end{center}
\caption{Automatic evaluation of multi-label classification models.}
\label{tab:result_f}
\end{table}

\section{Treatment Prioritization Model}

\subsection{Proposed Methods}

We develop a prioritization model of treatment on the datasets shown in (b) and (c) of Figure~\ref{fig:annotation}.
This model takes as input a list of text representing the orthodontic problems or a list of classes and outputs a list of treatment priority for each problem shown in parentheses in the lower part of Figure~\ref{tab:result_f}.

We convert each problem into vector representation using a natural language processing approach and perform learning-to-rank with them as features.
The following three methods are used to vectorize each problem.
\begin{itemize}
    \item BoW (Bag-of-Words): We construct feature vectors from the text of the problem. These vector representations have dimensions corresponding to the number of vocabulary in the training dataset, and each dimension has a value of $1$ if the corresponding word appears in the input text or $0$ it does not.
    \item OoK (One-of-K): We construct feature vectors from the labels that represent the problem. These vector representations have dimensions corresponding to numbers of labels appearing in the training dataset, and only one of the dimensions corresponding to the input class has a value of $1$ and the other dimensions have a value of $0$.
    \item USE (Universal Sentence Encoder): We construct feature vectors from the text of the problem using the cross-lingual version of universal sentence encoder~\cite{chidambaram-2018}. These are $512$-dimensional dense vectors.
\end{itemize}

For learning-to-rank, we used SVM-rank~\cite{joachims-2006} with a linear kernel, a standard toolkit.

\subsection{Experimental Settings}

Each sentence was divided into words using MeCab~\cite{kudo-2004} for the BoW model.
In this experiment, as the training dataset has $146$ vocabulary and $320$ classes, the feature vectors of each model are $146$ dimensions for BoW, $320$ dimensions for OoK, and $512$ dimensions for USE.
For a given set of features, we examined a hyper-parameter among $C \in \{1, 5, 10, 50, 100, 500, 1000, 5000\}$ on the validation dataset for SVM-rank.

Spearman's rank correlation coefficient is used to automatically evaluate the performance of each model.
When the correlation coefficient between the human ranking and the estimated ranking exceeds $0.4$, the estimation result of the model can be interpreted as having a positive correlation with human evaluation.

\subsection{Experimental Results}

The experimental results are shown in Table~\ref{tab:result_s}.
As each method has Spearman's rank correlation coefficient exceeding $0.4$, it can be interpreted that these estimation results have a positive correlation with human evaluation.
Compared to the BoW model, the OoK model is expected to be able to obtain feature vectors that reflect an annotator's expertise, so it is considered that high performance has been achieved.
Furthermore, the USE model achieves the highest performance because the dense vectors obtained by deep learning can represent rich information.

\begin{table}[t]
\begin{center}
\begin{tabular}{lc}\toprule
    & $\rho$ \\ \midrule
BoW & 0.513  \\
OoK & 0.566  \\
USE & 0.584  \\ \bottomrule
\end{tabular}
\end{center}
\caption{Automatic evaluation of treatment ranking models by Spearman's rho.}
\label{tab:result_s}
\end{table}

\section{Conclusion}
We work on the task of automatically designing a treatment plan from the findings included in the medical certificate written by the dentist.
To develop an artificial intelligence system that deals with free-form certificates written by dentists, we annotate the findings and utilized the natural language processing approach.
As a result of the experiment using 990 certificates, $0.585$ F1-score was achieved for the task of extracting orthodontic problems from findings, and $0.584$ correlation coefficient with the human ranking was achieved for the treatment prioritization task.

Our future work includes:
\begin{itemize}
    \item Fine-tuning on in-domain data of sentence encoder~\cite{devlin-2018}.
    \item Robust sentence encoding for incomplete sentences for multi-label classification model.
    \item Consideration of the findings for treatment prioritization.
    \item Text simplification from treatment protocol summaries to consent form documents.
\end{itemize}

\clearpage
\bibliographystyle{unsrt}

\begin{thebibliography}{9}
  \bibitem{takada-2017} Kenji Takada. Elements of Orthodontics. 2017.
  \bibitem{simswilliams-1986} J. H. Sims-Williams, I. D. Brown, A. Matthewman and C. D. Stephens. A Computer-Controlled Expert System for Orthodontic Advice. British Dental Journal, vol.163, no.5, pp.161--166, 1987.
  \bibitem{stephens-1996} C. D. Stephens, N. Mackin, J. H. Sims-Williams. The Development and Validation of an Orthodontic Expert System. British Journal of Orthodontics, vol.23, no.1, pp.1--9, 1996.
  \bibitem{chidambaram-2018} Muthuraman Chidambaram, Yinfei Yang, Daniel Cer, Steve Yuan, Yun-Hsuan Sung, Brian Strope and Ray Kurzweil. Learning Cross-Lingual Sentence Representations via a Multi-task Dual-Encoder Model. arXiv:1810.12836, 2018.
  \bibitem{tokui-2015} Seiya Tokui, Kenta Oono, Shohei Hido and Justin Clayton. Chainer: a Next-Generation Open Source Framework for Deep Learning. In Proceedings of Workshop on Machine Learning Systems in The Twenty-ninth Annual Conference on Neural Information Processing Systems, 2004.
  \bibitem{kudo-2004} Taku Kudo, Kaoru Yamamoto and Yuji Matsumoto. Applying Conditional Random Fields to Japanese Morphological Analysis. In Proceedings of the 2004 Conference on Empirical Methods in Natural Language Processing, pp.230--237, 2004.
  \bibitem{joachims-2006} Thorsten Joachims. Training Linear SVMs in Linear Time. In Proceedings of the ACM Conference on Knowledge Discovery and Data Mining, 2006.
  \bibitem{devlin-2018} Jacob Devlin, Ming-Wei Chang, Kenton Lee and Kristina Toutanova. BERT: Pre-training of Deep Bidirectional Transformers for Language Understanding. arXiv:1810.04805, 2018.
\end{thebibliography}

\end{document}